# Can professional translators identify machine-generated text?

**Michael Farrell**
IULM University
Milan
Italy
michael.farrell@iulm.it

## Abstract

This study investigates whether professional translators without prior specialized training can reliably identify short stories generated in Italian by artificial intelligence (AI). Sixty-nine translators took part in an in-person experiment, where they assessed three anonymized short stories — two written by ChatGPT-4o and one by a human author. For each story, participants rated the likelihood of AI authorship and provided justifications for their choices. While average results were inconclusive, a statistically significant subset (16.2%) successfully distinguished the synthetic texts from the human text, suggesting that their judgements were informed by analytical skill rather than chance. However, a nearly equal number misclassified the texts in the opposite direction, often relying on subjective impressions rather than objective markers, possibly reflecting a reader preference for AI-generated texts. Low burstiness and narrative contradiction emerged as the most reliable indicators of synthetic authorship, with unexpected calques, semantic loans and syntactic transfer from English also reported. In contrast, features such as grammatical accuracy and emotional tone frequently led to misclassification. These findings raise questions about the role and scope of synthetic-text editing in professional contexts.

## 1 Introduction

Today, authors working in a second language can circumvent the conventional approach of first writing in their native tongue and subsequently translating the text — whether via a human translator or machine translation (MT) — by crafting tailored prompts for generative artificial intelligence (GenAI). These prompts may be written in the author's first language, the target language or a mix of both, and provide a detailed description, structure or preliminary draft of the intended content. As a result, there is no source text document in the traditional sense (Farrell, 2025b).

Content produced in this way may subsequently be polished by a human synthetic-text editor, whose task is to make the text "more engaging" and give it "a more human voice" (Farrell, 2025a). This kind of editing requires a distinct skill set compared to post-editing, since the issues commonly found in synthetic text (SynT) — including repetitiveness, blandness, excessive wordiness, low burstiness (little variation in sentence length and structure) and superficial analysis — differ notably from the typical flaws seen in raw MT output (Dou et al., 2022; Farrell, 2025a). Unlike MT, where distortions may arise from source-text transfer among other factors, SynT is generated without an explicit source text. Most of the reported anomalies are found both in English and Italian SynT (Farrell, 2025b) and are most likely found in SynT in other languages too.

The rationale behind synthetic-text editing (STE) is based on the assumption that readers can, in fact, distinguish between AI-generated and human-authored texts. Similarly, the ability to recognize the hallmark traits of SynT is clearly an essential skill for effective STE.

Clark et al. (2021) reported that untrained, non-expert evaluators generally perform poorly at detecting machine-generated English text, and even after training, their success improved only slightly, reaching approximately 55%. In contrast, Dou et al. (2022) demonstrated that laypeople could differentiate between English SynT and human-authored text (HAT) when using an annotation framework called Scarecrow, which defines specific error types.

A more recent study by Farrell (2025b) found that postgraduate translation students, on the whole, struggle to identify SynT even after preparatory training: only two out of 23 participants achieved notable accuracy. However, given that the likelihood of two students succeeding purely by chance was calculated at around 10.32%, this outcome strongly suggests that their performance involved analytical reasoning rather than random guessing.

Hypothesizing that professional translators, with their language skills, may be better qualified for the task than postgraduate students, it was decided to conduct an experiment on one such cohort.

## 2 Aims

The principal objective of this experiment was:

- To determine whether professional translators, thanks to their experience in manipulating text, can effectively identify Italian SynT.

There were also several secondary aims:

- To determine whether age, gender, educational background, native language and years of experience as a translator have any effect on ability to identify Italian SynT.
- To have participants report the textual anomalies they find in the hopes these can reliably form the basis of training materials for Italian SynT detection and STE.
- To assess whether ChatGPT-4o can be prompted to write in the style of a particular Italian author.
- To shed light on the actual need for STE. If professional translators fail to consistently identify machine-generated text, it may indicate that Italian SynT is sufficiently human-like not to require any particular editing.

## 3 Method

Sixty-nine professional translators were brought together in person in Milan, Italy. Each received an envelope containing printed hardcopies of three unabridged, randomly ordered short stories of comparable length written in Italian. The titles were removed and replaced with geometric shapes (an oval, a hexagon, and a five-pointed star) to allow identification during analysis. These shapes were chosen since they lack a natural order and therefore do not suggest any obvious ranking.

Assuming an average reading speed of 200 words per minute, the estimated reading time for all three texts was approximately 24 minutes. However, since participants were also asked to provide written feedback, the session concluded only after the last participant had finished (after about 50 minutes).

The participants were informed that between zero and all three of the stories were machine-generated and between zero and all three were human-written. In reality, two stories were generated by ChatGPT-4o and one was written by a human author.

At the end of each story, the participants were asked to complete a form and assign a score from 0 to 10 to each text based on its likelihood of being artificial (0 = human; 10 = machine-generated; 5 = uncertain). Intermediate integer scores were also allowed. Additionally, participants were told to underline the portions of the text that informed their judgements, explain their conclusions, and note whether they believed they had read the story before or recognized the style of a specific author.

To prevent the participants from distinguishing the HAT by locating excerpts online, they were not allowed to use internet-enabled devices such as laptops or smartphones. The participants were also required to form their own opinions independently. Consequently, they were not allowed to consult with anyone, including fellow participants.

Each participant also completed a brief demographic questionnaire detailing their age, gender, first language, educational background and years of experience as a translator.

At the end of the experiment, the participants were instructed to place all the materials they received back into the envelope, seal it and return it to the researcher.

### 3.1 Preparatory training

In the previous experiment (Farrell, 2025b), only 2 of the 23 students (8.70%) were able to correctly identify the SynT after preparatory training, suggesting that the training provided was of limited effectiveness. Furthermore, half of the participants explicitly deemed the training insufficient. Moreover, the students identified most of the textual anomalies they were trained to look out for in both the HAT and SynT excerpts. In light of this, and to minimize the risk of misleading participants with poorly designed materials, no prior AI-detection training was provided to the translators in this study.

## 3.2 Statistical analysis

The probability that k participants distinguish the SynTs from the HAT correctly purely by chance can be calculated using the binomial probability formula[1]:

$$P(k) = \binom{n}{k} p^k (1-p)^{n-k}$$

Where n represents the total number of participants, and p is the probability that an individual participant guesses correctly. To calculate p, we note that the participants were told that between zero and all three stories were HATs and the others were SynTs. Therefore, there are four possible scenarios, ranging from "none of the texts is an HAT" to "all three texts are HATs."

Hence, the probability of a participant guessing that only one of the texts is written by a human is 1/4. If they correctly guess this, the probability of guessing which one it is without looking at the texts is 1/3, as there are three stories. Since these two guesses are independent, the combined probability of making both correctly is 1/4 * 1/3 = 1/12[2].

The non-categorical demographic data was transformed into categorical data by grouping values into bands containing approximately equal numbers of participants. For instance, age was divided into the following ranges: ≤40, 41–50, 51–55, 56–60, and >60. To put some order into the wide variety of qualifications reported, these were grouped into two categories: postgraduate and non-postgraduate. First language was analysed as Italian vs. other.

Contingency tables were then drawn up, with the rows representing whether participants succeeded in identifying the kind of text. For the two SynTs, a score of 7–10 points was considered successful identification, while a score of 0–6 indicated failure. For the HAT, success was defined as a score of 0–3 points, and failure, a score of 4–10. The 2×2 tables were analysed using Fisher's exact test, while the larger tables were analysed using the chi-squared test. The statistics calculators provided by Stangroom[3] were used to perform the analyses.

## 3.3 Texts used

The two AI-generated texts in this experiment were those most frequently misidentified as human in the previous postgraduate study (Farrell, 2025b). Full details of the prompts and the prompt engineering techniques used are available in the cited paper. The SynT marked with a hexagon corresponds to ST5 in the previous study, for which ChatGPT was asked to write in the style of Giorgio Faletti, while the one marked with a star corresponds to ST2, where no author was specified. The human-written text was once again Alberto Moravia's short story *L'incosciente* (The Reckless Man), from *Racconti romani* (Roman Tales, 1954), marked with an oval. It was selected primarily for its brevity and because it was written well before the advent of MT and GenAI, which ensures that these technologies could not have played any role in its creation.

Unlike in the previous experiment, the three short stories were not divided into consecutive short extracts but were presented in full to the participants. This protocol change was based on the reasoning that low burstiness and narrative contradiction — identified in the earlier study as potentially the most reliable indicators of AI-generated texts — are probably easier to detect over longer passages.

| Text | Kind | Length (words) | Mean human score | Standard deviation | AI Detector score |
|---|---|---|---|---|---|
| **Oval** | Human | 1807 | 4.78 | +/- 4.13 | 17% |
| **Star** | Synthetic | 1331 | 4.22 | +/- 4.08 | 83% |
| **Hexagon** | Synthetic | 1626 | 5.61 | +/- 3.77 | 94% |

**Table 1.** Assigned scores

[1] The Statology binomial distribution calculator was used: www.statology.org/binomial-distribution-calculator

[2] Given the nature of the experiment, some participants may have doubted that none of the texts were human-written. If the participants were guessing in the belief at least one of the texts was human-written, the chance of guessing that exactly one of the texts was is 1/3, giving a combined probability of 1/9. However, two participants believed all three texts were human-authored, while one thought all were AI-generated.

[3] https://www.socscistatistics.com/tests

All three texts (the two SynTs and the HAT) were analysed using the Plagramme AI detector[4] to determine whether any objectively measurable differences existed between them.

# 4 Results

## 4.1 Ability to identify SynT

The experiment involved 69 professional translators in two sessions on 10 and 19 December 2024. One participant was disqualified for not assigning scores to all three texts.

Given a ten-point scale where 0 = human, 10 = machine-generated, and 5 = uncertain, a participant may reasonably be considered to have successfully distinguished the HAT from the SynTs if they assign the HAT a score below 5, the two SynTs scores above 5, and the difference between the lowest score and each of the two highest scores is at least 4 points.

Of the 68 valid participants, 11 met this criterion. Raising the required score difference to 5 points still gives 10 qualifying participants. Using the method described above, the probability of at least 11 participants out of 68 succeeding purely by chance is approximately 2.45%[5]; even with a 5-point threshold, the probability remains low at 5.47%. These figures strongly suggest that the translators concerned used analytical skills rather than guess work when distinguishing between HAT and AI-generated texts, which is in contrast with the impression we get looking at the mean scores in Table 1.

It should be further noted that 9 participants came to the opposite conclusion: they assigned the HAT a score above 5 and the two SynTs scores below 5, again with at least a 4-point difference. Raising the difference threshold to 5 points does not change this number.

The near-equal numbers (11 vs. 9) help explain why the mean scores shown in Table 1 obscure the participants' actual ability to identify the HAT — the correct and incorrect evaluations tend to cancel each other out. However, if we consider all 20 participants who clearly found the HAT distinct from the others, regardless of direction, the result far exceeds what would be expected by chance.

## 4.2 Influence of demographic variables

Despite clear instructions, only 57 of the 68 valid participants completed the demographic questionnaire. Due to an oversight, the question on first language only appeared in the version used during the second session, on 19 December, resulting in just 26 responses: Italian (22), German (2), French (1), and Russian (1). When reclassified as Italian vs. non-Italian, a Fisher's Exact Test showed the proportions were representative of the experimental population according to the recruitment data, suggesting the language data can reasonably be considered to reflect the entire group.

No significant associations were found between ability to distinguish between SynT and HAT and education level, age group, first language or years of professional experience for any of the three short stories.

Regarding gender, Fisher's exact test for the hexagon text gave a p-value of 0.0269, indicating a statistically significant result at $p < .05$. This suggests that male participants were more successful at identifying the text as artificial. However, the finding is statistically fragile: only four of the 68 participants were actually male, reflecting the female dominance of the translation profession, and a single different male response would have rendered the result non-significant. For the other two texts, no significant gender-related differences were observed.

## 4.3 Reported textual features

Tables 2-4 in the *Appendix* summarize the reasons provided by participants who correctly or incorrectly identified the nature of each of the three texts, excluding uncertain responses (scores of 4–6). These reasons are examined in the *Discussion* section, and the most salient are laid out in the subsections below.

### 4.3.1 Influence of English

Several translators reported noticing influence from English in the SynTs, particularly in syntax and usage that diverged from standard Italian norms. Most of these criticisms concerned the hexagon text.

#### 4.3.1.1 Syntactic or pragmatic transfer

In Italian, it is common to drop possessive adjectives when the context is clear, especially with

[4] www.plagramme.com

[5] Even in the case of the "disbelief" described in footnote 2, the probability is still low at approximately 12.98%.

body parts (Serianni, 1988; Brunet, 1980). While the underlined possessive adjectives in the passage below from the hexagon text are not grammatically incorrect, omitting them would sound more natural:

*“La stanza, solitamente accogliente con i suoi mobili antichi e le pareti tappezzate di libri, ora sembrava stringersi attorno a lui. Si sedette alla sua scrivania, la penna in mano, la carta bianca davanti a lui. Ogni volta che cercava di iniziare a scrivere, la mano tremava leggermente.”*

As the fairly literal English translation below shows, such use of possessives sounds natural in English. However, the definite article in the final line, which is perfectly idiomatic in Italian, renders awkwardly in English, where a possessive adjective would be expected:

*"The room, usually welcoming with its antique furniture and walls lined with books, now seemed to close in on him. He sat at his desk, pen in hand and blank paper before him. Every time he tried to start writing, the hand trembled slightly."*

The use of more possessive adjectives than is natural in native Italian was observed in both of the AI-generated texts used in this experiment, as well as in the other SynTs used in the earlier postgraduate study (Farrell, 2025b).

Another participant noted syntactic patterns which suggested English influence, such as the use of present participles following commas.

#### 4.3.1.2 Orthotypographic transfer

One participant pointed out punctuation inconsistencies, particularly regarding the placement of commas relative to quotation marks in direct speech, which is handled differently in Italian (Treccani, n.d.) and in standard American English (University of Chicago, 2017).

#### 4.3.1.3 Semantic loans

One translator noted that certain word choices, such as *speculazioni* (speculations), *casuale* (casual) and *sfida* (challenge), appeared to be direct semantic loans from English, since they sounded unnatural in the context. These are semantic loans rather than calques, as they involve the importation of English semantic traits into existing Italian words (Pulcini, 2023).

#### 4.3.1.4 Discourse-level calques

Some respondents noted phrases where deeper patterns of meaning and agency seem to have been transferred literally from English into Italian, resulting in expressions that lacked idiomatic naturalness in Italian, such as:

*"Ogni volta che vedeva un’auto passare vicino a casa sua, temeva fosse venuta a prenderlo."*

This sentence may be translated as “Every time he saw a car passing by his house, he feared it/she was coming to get him.” In this case, however, *she* cannot be the correct translation since the protagonist fears a male pursuer, leaving *it* (the car) as the only logical rendition. While the resulting sentence is perfectly acceptable in English, the construction is less natural in Italian, where it is typically the driver rather than the vehicle itself that is understood to “come and get” someone.

## 4.4 Author or text recognized

### 4.4.1 Oval text (HAT)

The authors the participants mentioned, Italo Calvino, Elsa Morante, Leonardo Sciascia and Giorgio Scerbanenco were all active during the mid-to-late 20th century. A more general guess of "a 1970s author" also situates the writer in a period when Alberto Moravia was highly productive.

Most of the named writers, including the actual author, Moravia, are associated with literary styles grounded in realism, social critique or political engagement. Even Calvino, who later became more experimental, began with neorealism.

Although no participant correctly identified the author or recalled reading the story, it is interesting to note that Elsa Morante was, in fact, Moravia’s wife. However, their styles and themes differ significantly: Moravia's work is consistently realistic, while Morante's often takes on a dreamlike quality.

### 4.4.2 Hexagon text (SynT)

One reader thought the story might have been written by Spanish author Carlos Ruiz Zafón. In reality, ChatGPT-4o was prompted to write this text in the style of Italian writer Giorgio Faletti. Both Zafón and Faletti are known for their strong narrative drive and their use of mystery and psychological tension. While Faletti focuses on crime and high-stakes thrillers, Zafón explores literary mysteries with a gothic, atmospheric touch. The Italian setting (Asti), however, makes Zafón an unlikely match.

#### 4.4.3 Star text (SynT)

The authors mentioned, Carlo Cassola, Giorgio Bassani and Carlo Emilio Gadda, were contemporaries, mainly active from the 1940s to the 1960s. Cassola and Bassani specialized in different forms of realism, while Gadda produced experimental, linguistically complex narratives, quite removed from Moravia, or even from Cassola and Bassani. In fact, the linear style of the latter two aligns more closely with Moravia's.

The mention of "a female author from the 1980s" refers to a slightly later period. Interestingly, this was the only story one respondent believed they had read before.

## 5 Discussion

### 5.1 Ability to identify SynT

In the previous study involving postgraduate students, only two out of the 23 participants (8.70%) were able to correctly identify Italian SynTs (Farrell, 2025b). In the current experiment, eleven out of the 68 professional translators (16.2%) successfully identified Italian SynTs. If we also consider the nine translators who reversed the classification, mistaking SynTs for HATs and vice versa, we can say that a total of 20 participants (29.4%) perceived a difference between the two kinds of text.

However, this experiment is not directly comparable to the previous one due to significant differences in their protocols, most of which were designed to improve detection rates. The preparatory training, deemed insufficient and potentially responsible for spurious results, was omitted. Additionally, participants were presented with unabridged short stories rather than short excerpts, based on the assumption that certain textual anomalies would become more apparent over longer passages. Moreover, as a result of using longer texts, the number of stories was reduced from seven to three, although the selected SynTs were those previously found to be more difficult to identify.

### 5.2 Influence of demographic variables

The lack of any significant association between text identification ability and professional experience suggests that this skill is either acquired early on in one's translation career or is a form of sensitivity some individuals naturally possess and therefore had prior to entering the profession. If the latter is the case, then the improvement in results may be entirely attributable to the changes in the experimental protocol.

Perhaps the most surprising result from the demographic analysis was the lack of a statistically significant association between the participants' first language and their ability to identify AI-generated Italian text. One possible explanation is that, as professional translators working from Italian, the participants are used to reading Italian in a highly analytic way. Alternatively, AI detection may require a special language-independent skill.

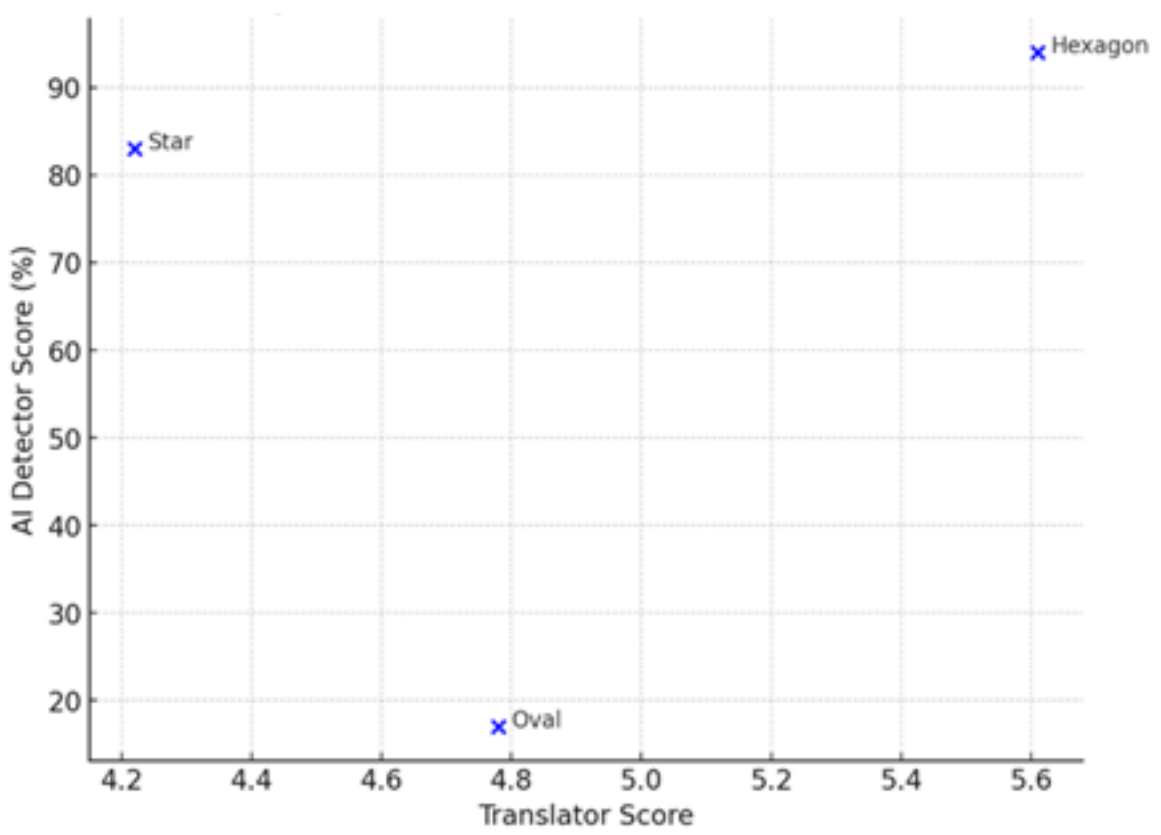

**Chart 1.** Scatterplot of scores assigned by the translators vs. AI detector scores

The scatterplot in Chart 1 shows there is no clear inverse or direct correlation between the scores assigned by the translators and AI detection scores in this small sample. Spearman's rho is 0.5 suggesting a moderate positive rank correlation between the translator's scores and AI detection scores. However, the high p-value (0.67) indicates that this correlation is not statistically significant with such a small sample size (n = 3).

### 5.3 Reliability of textual features

#### 5.3.1 Oval text

Those who correctly identified the text as HAT noted its lexical and stylistic richness, along with varied sentence structures. These traits are consistent with high burstiness, a feature of natural human writing. No instances of narrative contradiction were reported. This is also a hallmark of HAT since published authors typically work with editors and proofreaders who catch plot holes, timeline issues and inconsistencies before publication.

Conversely, those who misclassified the text as machine-generated cited frequent grammar and orthographic errors, such as misuse of the euphonic -d[6]. However, it is likely that Moravia deliberately used unconventional grammar to convey his characters' social and cultural backgrounds, thereby enhancing authenticity. Some participants claimed the story lacked depth or nuance and therefore must be SynT. This reveals a perceptual bias: they expected AI to lack depth and interpreted perceived dullness as machine authorship. Nevertheless, dull human writing exists, so equating boring with synthetic is logically flawed. There were also complaints about fragmented structure and weak logic, but such traits are common in experimental or stream-of-consciousness styles and are therefore unreliable indicators. In reality, neither style describes Moravia's writing. However, it is entirely possible that some participants disliked the story or Moravia's writing style so much that they rejected the idea it was human-written.

Rare or unexpected vocabulary, such as *grassazione* (armed robbery), *forbito* (eloquent), *impostare* (to post), *paltò* (coat), *cippo* (cippus), *rettifilo* (straight road) and *parabrise* (windscreen), divided opinion. Some viewed these as evidence of human creativity, others as a sign of machine-generated language, depending on whether the deviation was perceived as a literary device or an error. Had participants been allowed internet access (see *Methods*), they very probably would have drawn different conclusions.

### 5.3.2 Hexagon text

The participants who correctly identified the text as synthetic pointed to its repetition, lack of variety, overuse of names and formulaic phrasing and structure. These features reflect low burstiness, with lexical and syntactic patterns appearing flat and repetitive rather than clustered and variable as in human writing.

One clear instance of narrative contradiction was identified:

*"Quando arrivò alla porta principale, esitò solo un momento prima di bussare. Il suono del campanello risuonò freddo e distante."*

This passage translates as "When he reached the front door, he hesitated only a moment before knocking. The sound of the doorbell rang out cold and distant." The juxtaposition of knocking with the subsequent ringing of the doorbell constitutes a continuity error.

In addition, logical gaps and unjustified plot twists were reported. At one point, Mr Rogliano claims to know that Emilio's actions were influenced by another person. While this is actually true, the narrative provides no information that Mr Rogliano could plausibly have known to support such a conclusion.

In contrast, those who misidentified the text as human-written focused on its fluency and grammatical correctness, mistakenly equating these with humanness. However, others reported errors in prepositions and idioms, suggesting a halo effect where overall fluency masked specific flaws.

The text's simple or limited vocabulary was interpreted differently: some saw it as a lack of lexical range (hence SynT), while others viewed it as intentionally simple but effective (and thus HAT). Similarly, the restrained, linear style struck some as robotic, while others perceived it as literary or emotionally powerful. The same narrative was perceived as both generic and emotionally deep, suggesting subjective differing views on what constitutes depth.

### 5.3.3 Star text

The translators who correctly identified the text as AI-generated noted it lacked variety, rhythmic unpredictability and stylistic clustering, which are features that typically signal bursty, human-like writing. Although no explicit narrative contradictions were cited, some comments pointed to a kind of stylistic incongruity that may be considered a soft form of contradiction.

Those who incorrectly identified the text as human-written were misled by its smoothness and emotional cues, interpreting these as signs of high burstiness. In reality, one deceived participant actually remarked that the writing was structurally and lexically repetitive (e.g., *tornare*, *riusciva*[7]) without recognizing this as an indicator of synthetic authorship.

Reactions to tone varied: some found it robotic or modelled, others natural and engaging. The restrained or even-toned writing was interpreted either as tasteful human subtlety or soulless AI mimicry depending on the reader's expectations.

[6] The euphonic -d refers to the addition of the letter d to the conjunction *e* (and), the preposition *a* (to/at) and, in the past, the conjunction *o* (or) when they precede a word starting with the same vowel to ease pronunciation.

[7] To return, managed.

Although no specific examples were provided, several responses referred to contradictions pointing to shaky, overly subjective or biased reasoning. Some said the plot was schematic and underdeveloped while others spoke of credible emotional and narrative flow.

#### 5.3.4 Across all three texts

The participants often cited the same features (e.g., simplicity, tone, emotional content), but used them to support opposite conclusions, saying they were signs of either HAT or SynT. Similar patterns were observed by Clark et al. (2021), who found that evaluators often gave contradictory reasons for their assessments. Apart from clearer indicators like burstiness, narrative contradiction and calques from English, most judgements appeared to have been driven more by subjective expectations, interpretive heuristics and bias than by objective criteria.

### 5.4 Interpretation of English influence

While calques, semantic loans and syntactic transfer are frequently observed in MT, see for example Quinci & Pontrandolfo (2023) and Aranberri & Pascual (2025), to the best of the author's knowledge, this is the first time such features have been reported in monolingual output generated by a large language model rather than in a translated text.

There are at least three plausible explanations for this phenomenon. Firstly, it is widely acknowledged that a large proportion of ChatGPT's training data is in English, which may lead to the transfer of English syntactic and lexical patterns into outputs in other languages. This English preponderance has been estimated at around 92.1% by character count, 93.7% by document count and 92.6% by word count for GPT-3 (Zread, n.d.). Secondly, the training corpus may contain machine-translated material, allowing existing cross-linguistic influence to be reproduced and reinforced. Thirdly, the prompt used to generate the Italian text was written in English, which is a common practice in prompt engineering. This may have primed the model to adopt English-language conventions in its output.

One participant remarked that certain passages read as if written by a native English speaker rather than a native Italian[8]. This raises the question of whether texts generated by ChatGPT in other languages besides Italian also give the impression of being written by a non-native author.

### 5.5 Authorship and style imitation

Regarding the authors identified, it seems that the participants who were not entirely off track based their suggestions more on the story's content than on a clear recognition of stylistic features. Overall, the experiment offers no strong evidence either for or against ChatGPT-4o's ability to convincingly emulate a specific author's style.

### 5.6 Training and STE implications

As also emerged from the previous study involving postgraduate students, future training in SynT identification and for synthetic-text editors could focus on identifying narrative contradiction and assessing variability in syntactic structures and lexical distributions, commonly referred to as burstiness (Farrell, 2025b).

In this experiment, only two participants (2.94%) picked up on the knock-on-door/doorbell-rings contradiction. While far from subtle, it was evidently not easy to spot. One might have expected professional translators to perform better, but in effect, they were only asked to read the text, not to translate it. The deeper analysis required in translation may have facilitated detection, which may suggest a possible training approach.

Teaching the identification of burstiness, however, may also prove difficult in practice. This study further confirms that other textual features, including grammatical accuracy and correct spelling, are unreliable indicators for distinguishing between Italian SynT and HAT.

Calques, semantic loans and syntactic transfer are additional issues that synthetic-text editors could be trained to recognize and address where appropriate. In this regard, it is worth noting that Italian is considered one of the languages most open to English influence, particularly in terms of lexical borrowing (Pulcini, 2023). However, since these features also occur in machine-translated texts and low-quality human translations, their presence alone is not a reliable indicator of SynT.

The need for STE is based on three key assumptions.

[8] *"Alcuni tratti mi sembrano un calco dall'inglese [...] come se [...] fosse stato scritto da una persona di madrelingua inglese."*

Firstly, that readers can distinguish between SynT and HAT. The results of this study show that 16.2% of the professional translators who took part were able to correctly identify Italian SynTs without any special training.

Secondly, that readers prefer HAT. In this study, nine participants (13.2%) incorrectly assigned human authorship to AI-generated stories and vice versa, possibly indicating a preference for SynT, which they evidently perceived as more human-like. This interpretation is supported by the findings of Zhang and Gosline (2023) who observed that English advertising content generated by GenAI was rated higher in quality than content produced solely by human experts. Similarly, Porter and Machery (2024) showed that AI-generated poetry was often indistinguishable from human-written verse and was rated more favourably overall. These results suggest that further research into reader preferences would be valuable, using the short stories from this study as test material.

Thirdly, that the author or client is willing to invest the additional time and resources required for STE. This is most likely in high-stakes or accuracy-critical domains such as medicine, law or science, where logical consistency and precision are paramount.

At times, STE may function much like a preference for handmade over mass-produced goods: not always strictly necessary but driven by a desire for authenticity.

In some cases, the authors might perform STE themselves if their proficiency in the target language permits, perhaps using GPT to recheck edited passages for grammaticality or fluency. This may be particularly effective when English is the target language, since such errors are relatively rare in English GenAI output. However, when cultural adaptation or appropriateness is required, the ideal SynT editor should be bicultural or, at least, highly knowledgeable about the target culture.

### 5.7 Limitations

This experiment was limited to a small selection of texts of a similar kind in a single language. As a result, the findings and conclusions of this study may not be broadly generalizable. However, the hope of identifying textual anomalies that may form the basis of training materials for SynT detection and potential contribution to STE training course development outweigh these limitations.

A further limitation concerns the absence of participant variables such as experience with MT, GenAI and editing, as well as relevant technological training, which may have influenced the participants' ability to distinguish between HAT and SynT.

## 6 Conclusion

This study explored the ability of professional translators to distinguish between synthetic (AI-generated) and human-authored Italian texts. While some participants correctly identified the SynTs, a comparable number made inverse misclassifications and the majority outright misjudgements, indicating that fluency, emotional cues and grammatical correctness can mislead even seasoned linguists.

The results support the case for specialized training in STE, which focuses on narrative contradiction, burstiness and English influence rather than linguistic accuracy or style. However, the findings also challenge the notion that STE is universally necessary, particularly when machine-generated texts may already be fit for purpose.

Further research is recommended to explore user preferences for AI-generated texts and cross-linguistic generalizability.

## Acknowledgments

The research project reported in this paper has received a Small Grant for Research from Mediterranean Editors and Translators.

## Appendix

Summary tables (2-4) of reasons provided by participants who correctly or incorrectly identified the nature of each of the three texts, excluding uncertain responses (scores of 4–6).

| | **Correctly identified as HAT** | **Incorrectly identified as SynT** |
|---|---|---|
| **Score** | 0–3 points | 7–10 points |
| **Participants** | 30 | 27 |
| **Lexical choice** | Rich, refined vocabulary; rare or regional terms; idiomatic and original expressions | Made-up or awkward words; odd adjectives; perceived as non-Italian |
| **Burstiness, fluency and syntax** | Complex, varied syntax; fluid, structured narrative; dynamic pacing; varied sentence lengths; expressive punctuation | Awkward, list-like structure; disjointed or monotonous sentences |
| **Dialogues** | Natural, lively; plausible informal grammar or regionalisms | Flat or forced; overly scripted |
| **Style, voice and narration** | Fresh, metaphorical, personal; distinct authorial presence | Mechanical or impersonal; generic or vague perspective; lacks depth or personality |
| **Register and tone** | Intentional mix of high and colloquial; literary or culturally nuanced | Inconsistent or oversimplified; forced sophistication without context |
| **Cohesion and coherence** | Clear logic and flow; strong thematic and stylistic unity; meaningful cultural references | Fragmented structure; weak logic; hard to follow |
| **Grammatical accuracy** | Minor, genre-consistent errors (e.g., regionalisms) | Frequent, distracting mistakes; unnatural structure and punctuation |
| **Narrative contradiction** | None | None |

**Table 2.** Reasons given for oval text score

| | Correctly identified as SynT | Incorrectly identified as HAT |
|---|---|---|
| **Score** | 7–10 points | 0–3 points |
| **Participants** | 30 | 22 |
| **Lexical choice** | Repetitive (e.g., Emilio, *sfida*); clichés; limited synonyms; calques (see *Influence of English*) | Simple but expressive vocabulary; occasional originality |
| **Burstiness, fluency and syntax** | Mechanical phrasing; awkward structures; agreement errors; repetitive patterns and monotonous rhythm | Smooth, fluid sentences; good sentence flow enhanced engagement |
| **Dialogues** | Sparse, unidiomatic; inconsistent with narration | Matched narrative tone; added emotional and psychological depth |
| **Style, voice and narration** | Flat, linear, robotic; overuse of pronouns; emotionally dull; generic or automated perspective | Readable and human-like; strong authorial presence; emotionally resonant, even moving |
| **Register and tone** | Simplistic or ill-suited; tone felt clichéd or flat | Register consistent; tone aligned with emotion; some said it felt too good for a machine |
| **Cohesion and coherence** | Disjointed plot; more a string of images than a story | Coherent structure; psychological continuity noted |
| **Grammatical accuracy** | Mostly correct but included preposition and idiom errors | High grammatical accuracy; nothing stood out as incorrect |
| **Narrative contradiction** | One major continuity error; some logical gaps (see *Discussion*) | Emotionally and narratively consistent |

**Table 3.** Reasons given for hexagon text score

| | Correctly identified as SynT | Incorrectly identified as HAT |
|---|---|---|
| **Score** | 7–10 points | 0–3 points |
| **Participants** | 26 | 32 |
| **Lexical choice** | Basic, repetitive vocabulary; clichés; bland phrasing; no lexical creativity | Precise, idiomatic language; varied tenses; minimal repetition |
| **Burstiness, fluency and syntax** | Short, uniform sentences; dry, report-like tone; low syntactic variation; rigid paragraph and sentence structure; mechanical rhythm | Smooth, logical flow; good sentence structure and tense use; immersive feel; no awkward phrasing |
| **Dialogues** | Sparse or absent; added to impersonal tone | Not explicitly mentioned, but flow felt natural and emotionally expressive |
| **Style, voice and narration** | Flat, linear, didactic; emotionally bland; felt prompted or modelled; generic or artificial tone; described as *GPT-like* | Emotionally rich and natural; expressions felt genuine; highly engaging; emotional depth supported belief in human authorship |
| **Register and tone** | Impersonal and overstructured; emotionally flat; felt artificial | Balanced and engaging tone; aligned with character psychology and age |
| **Cohesion and coherence** | Schematic or underdeveloped plot; summarizing style | Clear narrative flow; credible emotional and event progression |
| **Grammatical accuracy** | Generally correct, but with stiff syntax and odd punctuation | Idiomatic and fluid; no major errors |
| **Narrative contradiction** | None | None |

**Table 4.** Reasons given for star text score